\documentclass[conference]{IEEEtran}
\IEEEoverridecommandlockouts
\usepackage{cite}
\usepackage{amsmath,amssymb,amsfonts}
\usepackage{algorithmic}
\usepackage{graphicx}
\usepackage{textcomp}
\usepackage{xcolor}
\def\BibTeX{{\rm B\kern-.05em{\sc i\kern-.025em b}\kern-.08em
    T\kern-.1667em\lower.7ex\hbox{E}\kern-.125emX}}

\usepackage{amsthm}
\usepackage{amsmath,amssymb}
\usepackage{bm}
\usepackage{bbm}

\usepackage{algorithm}
\usepackage{algorithmic}
\usepackage{graphicx}

\usepackage{mathtools}
\mathtoolsset{showonlyrefs,showmanualtags}
\usepackage{bbm}

\usepackage{multirow}
\usepackage{booktabs}

\usepackage{subcaption}

\usepackage{url}
\usepackage{hyperref}

\usepackage{lipsum}

\AtBeginDocument{
  \abovedisplayskip     =0.3\abovedisplayskip
  \abovedisplayshortskip=0.3\abovedisplayshortskip
  \belowdisplayskip     =0.3\belowdisplayskip
  \belowdisplayshortskip=0.3\belowdisplayshortskip}

\begin{document}

\title{Balancing Summarization and Change Detection\\in Graph Streams
}

\author{\IEEEauthorblockN{Shintaro Fukushima}
\IEEEauthorblockA{\textit{Social System PF Development Division} \\
\textit{Toyota Motor Corporation}\\
\textit{Department of Information Science and Technology} \\
\textit{The University of Tokyo}\\
Tokyo, Japan \\
sfukushim@gmail.com}
\and
\IEEEauthorblockN{Kenji Yamanishi}
\IEEEauthorblockA{\textit{Department of Information Science and Technology} \\
\textit{The University of Tokyo}\\
Tokyo, Japan \\
yamanishi@g.ecc.u-tokyo.ac.jp}
}

\maketitle

\begin{abstract}
This study addresses the issue of balancing graph summarization and graph change detection. 
Graph summarization compresses large-scale graphs into a smaller scale. 
However, the question remains: 
To what extent should the original graph be compressed? 
This problem is solved from the perspective of graph change detection, 
aiming to detect statistically significant changes using a stream of summary graphs. 
If the compression rate is extremely high, 
important changes can be ignored, 
whereas if the compression rate is extremely low, 
false alarms may increase with more memory. 
This implies that there is a trade-off between compression rate in graph summarization and accuracy in change detection. 
We propose a novel quantitative methodology to balance this trade-off to simultaneously realize reliable graph summarization and change detection. 
We introduce a probabilistic structure of hierarchical latent variable model into a graph, 
thereby designing a parameterized summary graph on the basis of the minimum description length principle. 
The parameter specifying the summary graph is then optimized so that the accuracy of change detection is guaranteed to suppress Type I error probability (probability of raising false alarms) to be less than a given confidence level. 
First, we provide a theoretical framework for connecting graph summarization with change detection. 
Then, we empirically demonstrate its effectiveness on synthetic and real datasets. 
\end{abstract}


\theoremstyle{definition}
\newtheorem{definition}{Definition}[section]
\newtheorem{theorem}{Theorem}[section]
\newtheorem{assumption}[theorem]{Assumption}
\newtheorem{proposition}[theorem]{Proposition}
\newtheorem{lemma}[theorem]{Lemma}
\newtheorem{corollary}[theorem]{Corollary}

\newtheorem{example}{Theorem}[section]

\let\Bbbk\relax


\renewcommand{\algorithmicrequire}{\textbf{Input:}}
\renewcommand{\algorithmicensure}{\textbf{Output:}}


\newcommand*{\MyDef}{\mathrm{def}}
\newcommand*{\eqdefU}{\ensuremath{\mathop{\overset{\MyDef}{=}}}}
\newcommand*{\eqdef}{\mathop{\overset{\MyDef}{\resizebox{\widthof{\eqdefU}}{\heightof{=}}{=}}}}
\newcommand\mydef{\mathrel{\overset{\makebox[0pt]{\mbox{\normalfont\tiny\sffamily def}}}{=}}}

\renewcommand{\vec}[1]{\boldsymbol{#1}}
\def\transpose#1{#1^{\top}}
\newcommand{\argmax}{\operatornamewithlimits{argmax}}
\newcommand{\argmin}{\operatornamewithlimits{argmin}}
\def\d#1{\mathrm{d}#1}

\makeatletter
\def\env@cases{%
  \let\@ifnextchar\new@ifnextchar
  \left\lbrace
  \def\arraystretch{0.8}%
  \array{l@{\quad}l@{}}
}

\thickmuskip=0.3\thickmuskip
\medmuskip=0.3\medmuskip
\thinmuskip=0.3\thinmuskip

\section{Introduction}

\subsection{Purpose of This Paper}
\label{subsection:purpose_of_this_paper}

This study addresses the issue of 
the extent to which a graph stream should be compressed
to detect important changes in it. 
Recently, 
it is not uncommon to observe large-scale graphs. 
Therefore, it is necessary to extract essential 
information from such massive graphs 
to detect the changes. 
\textit{Graph summarization} is a promising approach for  
summarizing important attributes or parts of a graph as well as a graph stream
~\cite{Liu2018}. 
The compressed graph thus 
obtained is 
called a \textit{summary graph}. 
The more compressed the original graphs are, 
the better it is in saving computational complexity and space complexity. 
Although 
graph summarization provides a concise representation,  
detecting the intrinsic changes in that stream accurately and reliably is complicated.

\begin{figure}[t]
\centering
\includegraphics[width=\linewidth]{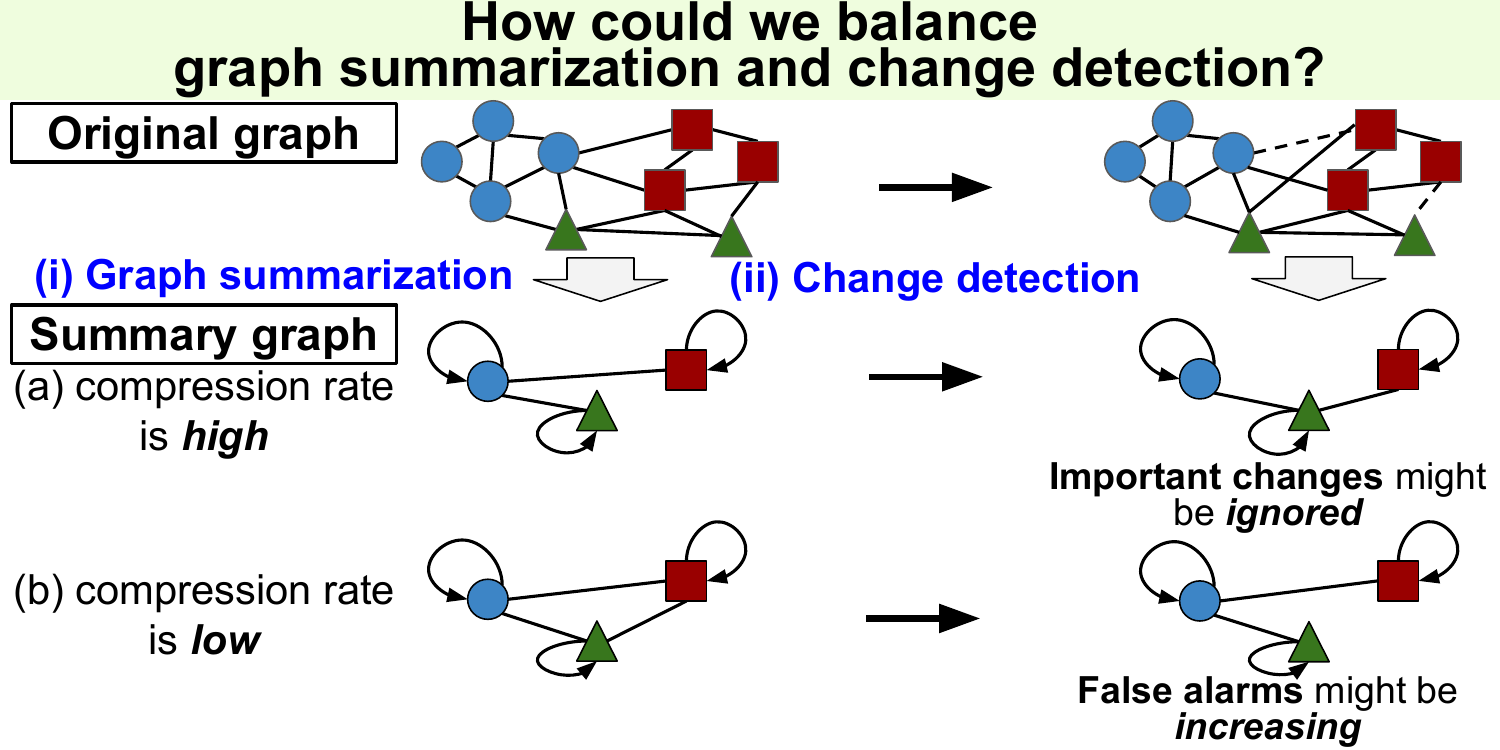}
\caption{
The issue of balancing graph summarization and change detection. 
An original graph is summarized into a summary graph.  
(a)~when the compression rate is too high, important changes might be ignored. 
(b)~when the compression rate is too low, false alarms of change might increase. 
}
\label{fig:conceptual_illustration_of_change_detection_with_graph_summarization}
\end{figure}

In Fig.~\ref{fig:conceptual_illustration_of_change_detection_with_graph_summarization}, 
original and summary graphs are shown at consecutive time points. 
Here, detecting the changes in the summary graphs is of interest. 
%
We note that changes detected using summary graphs depend on \textit{compression rate} of the summary graphs. 
Here, the compression rate refers to how compactly the original graph is summarized. 
When the compression rate is excessively high, 
that is, 
the original graph is compactly summarized, 
important changes may 
be overlooked. 
By contrast, when the compression rate is extremely low, that is, the original graph is poorly summarized, false alarms may increase because of the detailed changes detected, 
such as changes of connectivity between nodes. 
Therefore, 
there is a trade-off between compression rate of graph summarization 
and accuracy of graph change detection. 
Thus, the research question 
is summarized as follows: 
\textit{How can we balance summarization 
and change detection from 
a graph stream?} 
The purpose of this paper is to formulate  
this issue, 
and then to propose an algorithm 
from an information-theoretic perspective. 

\subsection{Related Work}
\label{subsection:related_work}


\subsubsection{Graph Summarization for Graph Streams}
\label{subsubsection:graph_summarization}

Many studies have proposed graph summarization algorithms for a static graph 
as well as 
a graph stream~\cite{Liu2018}. 
Graphscope~\cite{Sun2007} detects changes in encoding 
the cost of a graph segment.  
Com${}^{2}$~\cite{Araujo2014} finds temporal edge-labelled communities in a graph using tensor decomposition 
with the minimum description length~(MDL) principle~\cite{Yamanishi2023MDLBook}. 
TimeCrunch~\cite{Shah2015} finds graph structures with MDL 
using five vocabularies of graph structures 
and five temporal patterns.  
NetCondense~\cite{Adhikari2017} summarizes a dynamic network by aggregating nodes 
and time pairs. 
DSSG~\cite{Kapoor2020} identifies six atomic changes by combining the maximum entropy principle and MDL. 
However, none of these directly 
addresses 
the extent to which the original graphs should be 
summarized to detect significant changes 
in graph streams. 

\subsubsection{Change Detection in Graph Streams}
\label{subsubsection:change_detection}

Change detection in graph streams has been extensively explored~\cite{Akoglu2015,Ranshous2015}.  
Most of the previous studies addressed parameter changes in data distributions, 
such as 
(quasi-)distance-based methods~\cite{Hinkley1970,Basseville1993} 
and 
spectrum-based methods~\cite{Ide2004,Hirose2009,Akoglu2010}. 
DeltaCon~\cite{Koutra2016} detects anomalies based on (dis)similarity between successive graphs. 
An online probabilistic learning framework~\cite{Peel2015} 
combines a generalized hierarchical random graph model 
with a Bayesian hypothesis test.  
LAD~\cite{Huang2020laplacian} is a Laplacian-based change detection algorithm. 
HCDL~\cite{Fukushima2020} detects hierarchical changes  
in latent variable models with MDL.  


\subsection{Significance and Novelty}
\label{subsection:significance_and_novelty}


\begin{enumerate}
\item \textbf{Formulation for Balancing Summarization and Change Detection}: 
We formulate the issue  
of balancing summarization 
and change detection 
in a graph stream. 
Thus far, 
graph summarization and change detection 
have been developed independently. 
To the best of our knowledge, 
our study is the first attempt. 

\item \textbf{Novel Balancing Algorithm}: 
We propose a novel algorithm 
called 
the balancing summarization and change detection algorithm~(BSC). 
The key idea of BSC is to introduce a hierarchical latent variable model and to design a parameterized summary graph based on MDL~\cite{Yamanishi2023MDLBook}. 
With MDL, 
it is possible to decompose and combine code-lengths at different layers of variables of 
graphs, such as membership~(blocks) of nodes, 
(super)edges between the blocks, 
and edges in the original graph. 
The parameter specifying the summary graph is optimized so that the accuracy of change detection is guaranteed theoretically. 
This approach is 
related to \cite{Fukushima2020,Fukushima2021CN}. 
However, 
HCDL~\cite{Fukushima2020} only 
detects hierarchical changes in latent variable models for data streams 
and does not consider further compression of summary graphs, 
whereas 
the hierarchical latent variable probabilistic model~\cite{Fukushima2021CN} 
only summarizes a static graph 
and does not consider further compression nor change detection in an online setting. 

\item \textbf{Empirical Demonstration of the Effectiveness of the Proposed Algorithm}: 
We demonstrate the effectiveness of 
BSC empirically on synthetic and real datasets. We show that 
BSC takes an advantage over existing graph summarization algorithms,  
and 
is superior or comparable to change detection algorithms.  
\end{enumerate}


\section{Problem Setting}
\label{section:problemsetting}
This section presents a probabilistic framework for graph summarization and graph change detection prior to formulating 
the balancing issue within this framework.

\subsection{Summary Graph}
\label{subsection:summary_graph}

Let $G=(V, E)$ be a directed graph, where $V$ and $E$ denote a set of nodes and a set of edges, respectively. 
A summary graph $\bar{G}$ 
is defined as  
a triplet $\bar{G} = (S, U, \omega)$, 
where 
$S$ is a set of sets of nodes, 
$U$ is a set of sets of edges, 
and 
$\omega$ is an weight function~\cite{Lee2020}. 
$S$ is distinct and exhaustive subsets of  
$V$, i.e., 
$\bigcup_{A \in S} A = V$ and 
$\forall A \neq B \in S, A \cap B = \emptyset$. 
An element in $S$ is called a \textit{supernode}, and 
an element in $U$ is called a \textit{superedge}. 
We assume that each node in $V$ belongs to a single supernode in $S$. 
The weight function $\omega$ maps a superedge in $U$ onto a nonzero number. 
$U$ depends on how the graph is summarized, 
Typically, it is the total number of edges between supernodes~\cite{Lee2020}.

\subsection{Probabilistic Model of Graph Summarization}
Let $N$ be the number of nodes; $N=|V|$. 
Let ${\mathcal K}$ be a set of numbers of supernodes~(blocks) in summary graphs 
and 
$K$ be the number of 
supernodes belonging to $\mathcal{K}$. 
We introduce three random variables; 
$X=(x_{ij})\in \{0,1\}^{N\times N}$ 
is a random variable indicating the connections between nodes.
$Z=(z_{k})\in \{0,1\}^{N}$ 
is a random variable indicating the membership of each node corresponding to a supernode.
$Y=(y_{\ell m})\in \{0,1\}^{K\times K}$ 
is a random variable indicating a superedge connecting supernodes.  
Let $x,y,z$ denote the realizations of $X,Y,Z$, respectively; 
$Y$ and $Z$ are considered as latent variables.
We formulate the probabilistic model for a graph with latent variables as follows:
\begin{align}
p(X, Y, Z; \phi, K) &\mydef  
p(X|Y, Z; \eta, K) 
p(Y|Z; \rho, K)
p(Z; \xi, K), 
\label{eq:hierarchical_latent_variable_model}
\end{align}
where
$\phi = (\eta, \rho, \xi)$ 
is a set of parameters; 
$\eta \in [0, 1]^{K \times K}$, $\rho \in [0, 1]$ are parameters of the Bernoulli distributions, 
and 
$\xi = \transpose{(\xi_{1}, \dots, \xi_{K})} \in [0, 1]^{K}$ is the 
one of the categorical distribution, 
where $\sum_{k=1}^{K} \xi_{k} = 1$.  
The generative model is 
as follows:
\begin{align}
x_{ij} &\sim
  \mathrm{Bernoulli}
  (
    x_{ij} | y_{z_{i} z_{j}}, z_{i}, z_{j}; 
    \eta, k
  ), \nonumber \\
y_{z_{i} z_{j}} &\sim 
  \mathrm{Bernoulli}( 
    y_{z_{i} z_{j}} | z_{i}, z_{j}; 
    \rho, k
  ),  \nonumber \\
z_{i} &\sim 
  \mathrm{Categorical}(
    z_{i}; 
    \xi, k
  ).  
\label{eq:hierarchical_sbm}
\end{align}
The random binary variable  
$y = ( y_{\ell m} )$ 
indicates whether a superedge exists between the supernodes 
$\ell$ and $m$ in a summary graph. 
The latent variable model 
in \eqref{eq:hierarchical_sbm} can be thought of as 
an extension 
of the stochastic block model~(SBM).
\begin{definition}{(Summary graph with hierarchical latent variable model)~\cite{Fukushima2021CN}}
\label{def:summary_graph_with_latent_variable_probabilistic_model}
A summary graph $\bar{G}$ of 
a graph $G$ 
in \eqref{eq:hierarchical_sbm} 
is defined as 
$\bar{G} = (S, U, \eta, \rho, \xi)$. 
$S$ and $U$ are the same as those in  
Sec.~\ref{subsection:summary_graph}, 
and 
$\eta$, $\rho$, and $\xi$ are parameters in \eqref{eq:hierarchical_sbm}. 
\end{definition}


\subsection{Balancing of Summarization and Change Detection}
Suppose that 
we obtain a sequence of graphs $G_{1},G_{2},\dots .$
Then, 
let us consider the following two issues: 1) graph summarization and 2) graph change detection.

Graph summarization 
obtains a summary graph $\bar{G}_{t}$ 
of an original graph $G_{t}$ at each time $t$. 
Based on the MDL principle~\cite{Yamanishi2023MDLBook}, 
the goodness of the summary graph is evaluated 
with code-length. 
That is, 
a summary graph should be chosen so that the total code-length 
for encoding the original graph as well as the summary graph is as short as possible. 
Let $\lambda \in \mathbb{R}^{+}$, 
given a graph $G_{t}$, find a summary graph such that,
\begin{equation}\label{eq_summary}
\min _{\bar{G}_{t}}\{L(G_{t}|\bar{G}_{t})+L_{\lambda}(\bar{G}_{t})\},
\end{equation}
where $L(G_{t}|\bar{G}_{t})$ is 
the code-length 
of $G_{t}$ for a given $\bar{G}_{t}$, 
and 
$L_{\lambda}(\bar{G}_{t})$ is the one of $\bar{G}_{t}$.  
We design $\lambda$ so that the compression rate would be controlled by $\lambda$, 
as explained in Sec.~\ref{subsection:change_detection}.

Graph change detection 
determines whether an intrinsic change in a sequence of summary graphs occurs 
at $t$. 
We employ the MDL principle again to measure how likely an intrinsic change happens, 
in terms of code-length. 
More specifically, 
with the same parameter $\lambda$ as in graph summarization, 
at $t$, 
we consider the following statistic:
\begin{equation}\label{eq_MDLCS}
\Phi _{t}=L_{\lambda}(\bar{G}_{t}\oplus \bar{G}_{t-1})-
\{L_{\lambda}(\bar{G}_{t})+L_{\lambda}(\bar{G}_{t-1})\}, 
\end{equation}
where $L_{\lambda}(\bar{G}_{t}\oplus \bar{G}_{t-1})$ is the code-length for the concatenation of $\bar{G}_{t}$ and $\bar{G}_{t-1}$, 
letting them have an identical probabilistic structure. 
The quantity~\eqref{eq_MDLCS} measures the degree of change 
in the reduction of the total code-length 
by separating the graph sequence 
at $t$. 
We refer to \eqref{eq_MDLCS} as the \textit{MDL change statistic}~\cite{Yamanishi2018,Yamanishi2016}.
Let $\epsilon_{\lambda}(>0)$ be a decision parameter dependent on $\lambda$. 
We determine that an intrinsic change occurs 
at $t$ 
if
\begin{equation}\label{eq_decision}
    \Phi_{t}>\epsilon _{\lambda},
\end{equation}
otherwise, we determine that a change does not occur.
We call this the {\em MDL change test}. 
$\epsilon _{\lambda}$ should be chosen so that the reliability of change detection is theoretically guaranteed 
as in Sec.~\ref{subsection:change_detection}.

It should be noted that an intrinsic change is distinguished from a change in appearance. 
The former refers to a statistically meaningful change, 
whereas the latter refers to a change of details, 
including the values of the parameters and the number of edges. 
Therefore, a change in the appearance does not necessarily indicate an intrinsic change.

The issues of graph summarization and 
change detection are 
closely related 
through 
$\lambda$. 
We may design $\lambda$ so that the larger $\lambda$ is, 
that is, 
the higher the compression rate is, 
the smaller $\epsilon _{\lambda}$ is. 
Eventually, 
the threshold for change detection can be further
relaxed. 
The $\lambda$ having such a property, 
is 
called the \textit{balancing parameter}. 
As the 
parameter 
varies, 
the concern is mainly 
on how compactly the original graph can be summarized 
to make the performance of change detection 
in the summary graph sequence as high as possible. 
This is our main concern for 
the balancing issue. 
The concrete design of 
$\lambda$ and code-length 
for \eqref{eq_summary} and \eqref{eq_MDLCS} are 
described in 
Sec.~\ref{section:proposed_method}.

\section{The Proposed Method: The BSC Algorithm}
\label{section:proposed_method}

This section presents the overall flow of BSC, 
focusing on the basic idea and resulting formulas. 
Detailed formulation and derivations are provided in 
the supplementary material\footnote{\url{https://drive.google.com/file/d/17pjbYFndePRcoRwG\_OidOqTUBbND3vt4}}. 

\subsection{Code-length for Graph Summarization}




We formulate the two code-lengths in \eqref{eq_summary}, where $L(G_{t} | \bar{G}_{t})$ is the code-length of the original graph $G_{t}$ for a given summary graph $\bar{G}_{t}$, 
and $L_{\lambda}(\bar{G}_{t})$ is the code-length of $\bar{G}_{t}$. 

First, 
we consider 
$L_{\lambda}(\bar{G}_{t})$. 
The summary graph $\bar{G}_{t}$ is specified by the latent variables $y_{t}$, $z_{t}$, 
and the number of blocks $k_{t}$. 
Therefore, it is necessary to consider the code-length required for encoding $y_{t}$ and $z_{t}$. 
We hierarchically encode 
(i) the number of groups $k_{t}$ in the summary graph, 
(ii) $z_{t}$, and 
(iii) $y_{t}$ given $z_{t}$, 
and then sum up the three code-lengths. 
Specifically, 
$L_{\lambda}(\bar{G}_{t})$ is defined as 
\begin{align}
L_{\lambda}(\bar{G}_{t})
&\mydef 
L_{\mathbb{N}}(k_{t}) 
+
L_{\mathrm{NML}}(z_{t}; k_{t}) 
+ 
L_{\mathrm{LNML}}(y_{t}|z_{t}; \lambda, k_{t}), 
\label{eq:def_codelength_summary_graph}
\end{align} 
where 
the three terms 
are code-lengths of (i), (ii), and (iii), respectively. 
$L_{\mathbb{N}}$ is the optimal 
code-length for positive integers
~\cite{Rissanen1978}: 
$L_{\mathbb{N}}(k) = 2.865 + \log{k} + \log{\log{k}} + \dots$, 
where the sum is taken over positive terms. 
$L_{\mathrm{NML}}$ is the {\em normalized maximum likelihood} (NML) code-length~\cite{Rissanen1978} and $L_{\mathrm{LNML}}$ is  the {\em luckiness NML}~(LNML) code-length~\cite{Grunwald2007}. 
NML code-length achieves the minimax regret~\cite{Yamanishi2023MDLBook} 
and is defined as
\begin{align}
L_{\mathrm{NML}}(z; k)
&\mydef 
-\log{ 
     p(z; \hat{\xi}(z), k)
 }
+\log{ \sum_{z'} p(z'; \hat{\xi}(z'), k)
 }  \\
&= -\sum_{k'} n_{k'} \log{ \frac{ n_{k'} }{N} } 
   +\log{ C_{Z}(k) }, 
\end{align}
where $\hat{\xi}(z)$ is 
the maximum likelihood estimator~(MLE) given $z$, 
$n_{k}$ is the number of nodes 
in the $k$-th block,  and 
$\log{ C_{Z}(k) }$ is the {\em parametric complexity} 
for the categorical distribution: 
\begin{align}\label{pc2}
\log{C_{Z}(k)}
= \log{ \sum_{z'} 
    p(z'; \hat{\xi}(z'), k)
  }. 
\end{align}
$C_{Z}(k)$ is computable with $O(N+k)$ 
using a recursive formula ~\cite{Kontkanen2007}. 
LNML code-length is defined as follows: 
\begin{align}
L_{\mathrm{LNML}}(y | z; \lambda, k) 
&\mydef 
-\log{ 
      p(y|z; \hat{\rho}(y, z), k) 
      \pi(\hat{\rho}(y, z); \lambda) 
 }  \\
&\quad 
 +\log{ C_{Y|Z}(k, \lambda)}, 
\label{eq:def_LNML}
\end{align}
where $\pi(\rho; \lambda)$ denotes the prior distribution of $\rho$. 
We set $\pi(\rho; \lambda)$ to 
the beta distribution: 
$\pi(\rho; \lambda) 
= 
\frac{ 
  \Gamma(a+b+\lambda)
}{ 
  \Gamma(a)
  \Gamma(b+\lambda)
}
\rho^{a-1}
(1-\rho)^{b+\lambda-1}$, 
where $\Gamma$ denotes the gamma distribution. 
The second term in \eqref{eq:def_LNML} is also referred to as the parametric complexity: 
\begin{align}\label{pc1}
\log{ C_{Y|Z} (k, \lambda) }
&\mydef 
\log{ 
  \sum_{y'} 
  p(y'|z; \hat{\rho}(y', z), k) 
  \pi(\hat{\rho}(y', z); \lambda)
}. 
\end{align}

Next, we consider 
$L(G_{t} | \bar{G}_{t})$. 
The original graph $G_{t}$ is specified by $x_{t}$ for given $y_{t}$, $z_{t}$, and $k_{t}$. 
Let $y_{t} = (y_{\ell m})$, where $y_{\ell m}$ denotes 
each superedge in the summary graph. 
Then, we consider the following two cases: 
(i) $y_{\ell m} = 1$ 
(ii) $y_{\ell m} = 0$. 
For case (i), the hierarchical structure of $x_{t}$, $y_{t}$, $z_{t}$, and $k_{t}$ in  \eqref{eq:hierarchical_sbm} 
holds.  
The code-length required for encoding $x_{\ell m}^{+} = \{ x_{ij}: y_{\ell m} = 1, z_{i}=\ell, z_{j}=m \} \subset x_{t}$ for given $y_{t}$, $z_{t}$, and $k_{t}$ is NML code-length 
for the Bernoulli distribution:
\begin{align}
&L_{\mathrm{NML}}(x_{\ell m}^{+} | y_{\ell m}=1, z_{t}; k_{t}) = \\
& 
-
  \log{ 
      \prod_{(i, j); z_{i} =\ell, z_{j} = m}
       p(x_{ij} | y_{\ell m}=1, 
       z_{i}=\ell, 
       z_{j}=m; 
       k_{t})
  }  \\
&
+ \log{
        \sum_{ 
            \{ x'_{ij} \}
        }
        \prod_{
            (i, j); z_{i}=\ell, z_{j} =m 
        }
        p(x'_{ij} | y_{\ell  m}=1, 
        z_{i}=\ell, 
        z_{j}=m;
        k_{t})
  }. 
\label{eq:nml_codelength_for_existing_superedge}
\end{align}
The second term in 
\eqref{eq:nml_codelength_for_existing_superedge} 
means the parametric complexity 
of the Bernoulli distribution. 

Next, 
for case (ii), 
the hierarchical structure of $x_{t}$ for given $y_{t}$, $z_{t}$, and $k_{t}$ does not hold because the corresponding superedge does not exist. Therefore, we adopt a different encoding scheme. 
Let us denote 
$x_{\ell m}^{-} = 
 \{
   x_{ij}: 
   y_{\ell m} = 0, 
   z_{i} = \ell, 
   z_{j} = m 
 \}
 \subset x_{t}$. 
The key concept is to encode the total possible number of edges in the original graph and the number of configurations for existing edges. 
The code-length required for encoding 
it is so-called the \textit{counting} code-length~\cite{Grunwald2007}: 
\begin{align}
\log{ (n_{\ell m} + 1) } 
 +
 \log{ 
   \begin{pmatrix}
   n_{\ell m} \\
   n_{\ell m}^{+}
   \end{pmatrix}
 }, 
\label{eq:counting_codelength_for_no_superedge}
\end{align}
where 
$n_{\ell m}$ 
is the total possible number of edges 
in the original graph 
between supernodes 
$\ell$ and $m$, 
and 
$n_{\ell m}^{+}$ 
is the 
one of existing edges. 

By combining \eqref{eq:nml_codelength_for_existing_superedge} 
and \eqref{eq:counting_codelength_for_no_superedge}, 
we formulate $L(G_{t} | \bar{G}_{t})$: 
\begin{align}
L(G_{t} | \bar{G}_{t})
&\mydef 
L(x_{t} | y_{t}, z_{t}; k_{t}) \\
&\mydef 
\sum_{\ell, m} 
\biggl\{
  y_{\ell m} 
  L_{\mathrm{NML}}(x_{\ell  m}^{+} | y_{\ell m} = 1, z_{t}; k_{t})  \\
&
  +(1 - y_{\ell m})
  \left(
    \log{ (n_{\ell m} + 1) } 
    + 
    \log{
      \begin{pmatrix}
      n_{\ell m} \\
      n_{\ell m}^{+}
      \end{pmatrix}
    }
  \right)
\biggr\}. 
\label{eq:code-length_for_encoding_x}
\end{align}

The optimization problem of estimating $\bar{G}_{t}$ is converted into 
\begin{align}
&\min_{\bar{G}_{t}}
\left\{ 
  L(G_{t} | \bar{G}_{t}) 
  +
  L_{\lambda} (\bar{G}_{t})
\right\} \\
&=
 \min_{k_{t}, y_{t}, z_{t}} 
 \Bigl\{
   L(x_{t} | y_{t}, z_{t}; k_{t})
   +
   L_{\mathbb{N}}(k_{t})
   +
   L_{\mathrm{NML}}(z_{t}; k_{t}) \\
& \quad \quad \quad \quad \quad 
   +
   L_{\mathrm{LNML}}
   (y_{t} | z_{t}; \lambda, k_{t}) 
 \Bigr\}. 
\label{eq:optimization_problem_for_estimatinig_summary_graph}
\end{align}
Given a set of the numbers of supernodes $\mathcal{K}$, 
for each 
$k_{t} \in \mathcal{K}$, 
we 
obtain $\hat{z}_{t}$ 
by estimating  
the block structures 
and 
parameter 
$\hat{\xi}_{t}$ of SBM. 
We 
determine 
$y_{\ell m}=1$ 
if NML code-length 
for encoding $x_{\ell m}^{+}$ 
is shorter than 
the counting code-length in 
\eqref{eq:code-length_for_encoding_x}, 
and 
$y_{\ell m}=0$ 
otherwise. 
Finally, 
we 
sum up the four code-lengths, 
and then select 
$\hat{k}_{t} \in \mathcal{K}$ 
with the minimum code-length. 

\subsection{Change Detection}
\label{subsection:change_detection}



We consider the formulation of \eqref{eq_MDLCS} and the threshold parameter $\epsilon_{\lambda}$ in \eqref{eq_decision}. 
First, 
$L_{\lambda}( \bar{G}_{t} \oplus \bar{G}_{t-1})$ in \eqref{eq_MDLCS} is defined as 
\begin{align}
&L_{\lambda}(\bar{G}_{t} \oplus \bar{G}_{t-1}) 
\mydef  \\
&\quad 
L_{\mathbb{N}}(k^{\oplus}) 
+
L_{\mathrm{NML}}(z_{t}; k^{\oplus})
+
L_{\mathrm{LNML}}(y_{t} | z_{t}; \lambda , k^{\oplus}) \\
&\quad 
+L_{\mathbb{N}}
(k^{\oplus})
+
L_{\mathrm{NML}}(z_{t-1}; k^{\oplus})  
+
L_{\mathrm{LNML}}(y_{t-1} | z_{t-1}; \lambda , k^{\oplus}), 
\end{align}
where $k^{\oplus}$ denotes the optimal number of blocks. 
The first three terms refer to the code-lengths 
to encode 
$\bar{G}_{t}$ 
and 
the last three terms refer to the 
ones 
to encode 
$\bar{G}_{t-1}$. 

Next, we consider the threshold determination for the change score $\Phi_{t}$, 
to raise an alarm for a change point in hypothesis testing. 
The evaluation metrics are Type I and II error probabilities; 
Type I error probability leads to a false alarm, 
whereas 
Type II 
one 
leads to overlooking a change point. 
We provide the upper bounds on Type I and II error probabilities in 
the supplementary material. 
Focusing on Type I error probability, a threshold $\epsilon_{\lambda}$ is specified so  that Type I error probability is upper-bounded by a confidence parameter $\delta$. 
Based on the upper bound of Type I error probability,  
we have the following theorem for the lower bound of $\epsilon_{\lambda}$. 

\begin{figure}[tb]
\centering
\includegraphics[width=\linewidth]{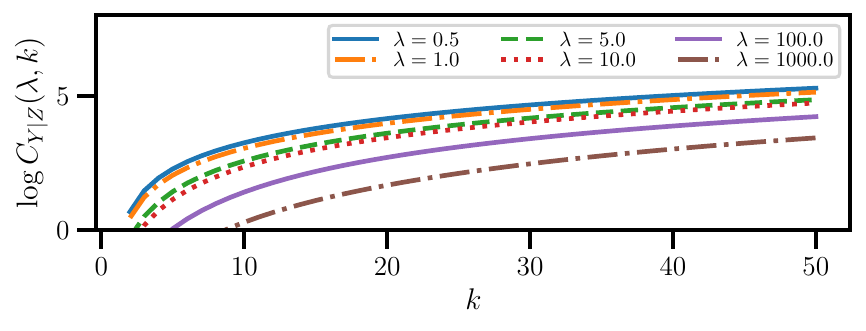}
\caption{
The dependence of the parametric complexity $\log{C_{Y|Z}(\lambda, k)}$ 
on 
$k$ and
$\lambda$. 
The parameters of the beta distribution are set to $(a, b) = (0.5, 0.5)$. 
}
\vspace{-0.5cm}
\label{figure:the_parametric_complexity_with_penalty_parameter_and_k}
\end{figure}

\begin{theorem}
Let $\epsilon_{\lambda}$ be the threshold for the MDL change test. 
For a balancing parameter $\lambda \in \mathbb{R}^{+}$, 
and confidence parameter $0<\delta <1$, 
$\epsilon_{\lambda}$ satisfies 
the following inequality: 
\begin{align}
\quad \quad 
\epsilon_{\lambda}
\geq 
\log{
    C_{Y|Z}(\lambda, k_{0}) 
    C_{Z}(k_{0}) 
}
+
\frac{ 
  L_{\mathbb{N}}(k_{0})
  -
  \log{\delta}
}{
  2
}, 
\label{eq:inequality_between_confidence_and_threshold}
\end{align}
where $k_{0}$ is the true number of blocks. 
Type I error probability does not exceed $\delta$. 
Here,
$\log{ C_{Y|Z}(\lambda, k_{0}) }$ 
and 
$\log{ C_{Z}(k_{0}) }$ 
are the parametric complexities in \eqref{pc1} and \eqref{pc2}, respectively. 
\end{theorem} 
The relationship between $\lambda$ 
and $\epsilon_{\lambda}$ represents a trade-off between compression rate of graph summarization and 
accuracy of change detection. 
Fig.~\ref{figure:the_parametric_complexity_with_penalty_parameter_and_k} illustrates the dependence of 
the parametric complexity 
on $k$ and $\lambda$. 
For a fixed $k$, 
the larger $\lambda$ is~(a high compression rate), 
the smaller 
the parametric complexity is, 
which leads to 
lower  
$\epsilon_{\lambda}$. 
In fact, 
it is not enough to evaluate only Type I error probability.  
In Sec.~\ref{section:experiments}, we empirically evaluate the
relationship between $\lambda$ 
and Type I / II error probabilities. 

\subsection{Overall Flow of BSC Algorithm}
\label{subsection:overall_flow_of_bsc_algorithm}


In summary, the overall flow of BSC is presented in Algorithm~\ref{algorithm:bsc}. 
The computational cost of BSC is evaluated as follows: 
for a fixed $k \in \mathcal{K}$, 
the computational cost of inference for SBM depends on the algorithm used. 
In this study, 
we use the MCMC-based  method~\cite{Peixoto2014};  
the computational cost is $O(N\log^{2}{N})$. Then, the computational cost for calculating the code-lengths in \eqref{eq:optimization_problem_for_estimatinig_summary_graph} 
is 
$O(N+k)$ for $L_{\mathrm{NML}}(z_{t}; k_{t})$ 
and 
at most 
$O(|E|)$ for $L(x_{t} | y_{t}, z_{t}; k_{t})$.
Therefore, the total computational cost for fixed $k$ is 
$O(N\log^{2}{N} + k + |E|)$. 
In total, the computational cost for the set $\mathcal{K}$ 
and 
the total time steps $T$ 
in a graph stream 
is 
$O(T(|\mathcal{K}|  (N\log^{2}{N} + |E|) + \sum_{k \in \mathcal{K}} k))$. 

\begin{algorithm}[tb]
\caption{Balancing Summarization and Change Detection Algorithm~(BSC)}
\label{algorithm:bsc}
\begin{small}
\begin{algorithmic}[1]
\REQUIRE $\delta$
: a confidence parameter 
to control Type I error probability,  
$\lambda$: penalty parameter ~($\lambda > 0$),  
$\mathcal{K}$: a set of the numbers of supernodes in summary graphs. 
\ENSURE
\FOR {$t=1, \dots$}
  \STATE Receive a graph $G_{t}$\,($=x_{t}$). 
  \STATE Estimate the number of blocks and latent variables 
  $(\hat{k}_{t}, 
    \hat{y}_{t},  
    \hat{z}_{t}
   )$ 
  according to \eqref{eq:optimization_problem_for_estimatinig_summary_graph} 
  with 
  $x_{t}$, 
  where $\hat{k}_{t}$ is selected 
  among $\mathcal{K}$.  
  \STATE Calculate the change statistics $\Phi_{t}$ 
  in \eqref{eq_MDLCS}. 
  \STATE Calculate the threshold 
  $\epsilon_{\lambda}$ 
  as the minimum of the inequality 
  in \eqref{eq:inequality_between_confidence_and_threshold}. 
  \IF {$\Phi_{t} > \epsilon_{\lambda} $}
    \STATE Raise an alarm of change 
    for summary graph. 
  \ENDIF
\ENDFOR
\end{algorithmic}
\end{small}
\end{algorithm}

\section{Experiments}
\label{section:experiments}

This section demonstrates the effectiveness of BSC 
through experiments. 
All experiments were performed on MacOS 10.15.7 with Intel Core i7 
and 16GB memory. 
The source code is available at 
\url{https://github.com/s-fuku/bsc}. 

\subsection{Synthetic Dataset}
\subsubsection{Data}

For $t=1, \dots, 30$, 
we generated edges $x_{t}$, 
where the number of nodes was set to $N=200$. 
First, 
$x_{1}$ was drawn from 
SBM~\cite{Snijders1997} 
with the parameters $\theta_{x}$, $\theta_{z}$, and model $k$; 
$\theta_{x}$ is the parameter in the Bernoulli distribution for edges 
and 
$\theta_{z}$ is the one 
in the categorical distribution 
for the blocks~(supernodes) of nodes. 
$\theta_{x}^{(1)}$ and $\theta_{z}^{(1)}$ were sampled 
from the beta distribution~($a=b=0.5$) 
and the Dirichlet distribution~($\alpha=1$), 
respectively. 
From $t=2$ to $14$, 
some edges were regenerated according to $\theta_{x}^{(1)}$ 
for each combination of supernodes with probability $0.05$. 
At $t=15$, 
each element in 
$\theta_{x}^{(1)}$ changed 
to 
$\theta_{x}^{(2)} = 
\max(
  \tau, 
  \min(
    \theta_{x}^{(1)} + u, 1 - \tau))$, 
where $u$ is drawn from the uniform distribution between $-0.1$ and $0.1$, 
and 
$\tau=0.001$. 
From $t=16$ to $30$, 
some edges were regenerated in the same way as before. 
We set 
$\delta=0.05$, 
$k \in \{ 2$, $3$, $4 \}$, 
and 
$\lambda \in \{ 0.5$, $1$, $5 \}$. 

\subsubsection{Algorithms for Comparison} 
\label{subsubsection:algorithms_for_comparison}
We chose 
the following two representative graph summarization algorithms with MDL 
and two graph change detection ones  
with the highest detection accuracy and 
relatively light-weight computation. 
%
\begin{itemize}
  \setlength{\parskip}{0cm}
  \setlength{\itemsep}{0cm}
    \item TimeCrunch\footnote{\url{https://github.com/GraphCompressionProject/TimeCrunch}}~\cite{Shah2015}  
    encodes 
    model and data, 
    where the model refers to 
    five connectivity structures 
    and 
    five temporal occurrences. 
    We calculated the code-length for the model as a summary graph at each time and used the absolute difference of it as a change score. 
    We chose the top $m \in \{ 10, 50 \}$ structures from the candidate set. 
    \item DSSG\footnote{\url{https://github.com/skkapoor/MiningSubjectiveSubgraphPatterns}}~\cite{Kapoor2020}  
    outputs 
    subgraph structures 
    as a summary graph 
    formed with 
    six atomic changes. 
    We defined the sum of \textit{information context}s 
    of all atomic changes at a time point 
    as a change score. 
    We changed $q$ 
    among $\{ 0.1$, $0.5 \}$, 
    where $q$ is expected probability of a random node in the summary graph. 
    \item  DeltaCon\footnote{\url{https://web.eecs.umich.edu/~dkoutra/CODE/deltacon.zip}}~\cite{Koutra2016} 
    calculates 
    similarity of 
    consecutive 
    pair of graphs. 
    We defined the change score 
    as $1-$ similarity. 
    \item Eigenspace-based algorithm~\cite{Ide2004} 
    outputs an anomaly score at each time 
    and we used it as a change score. 
    We set the window size $w=2$. 
\end{itemize}

\subsubsection{Evaluation Metrics}
We evaluated 
algorithms 
in terms of accuracy of change detection and compression rate of the summary graphs. 
We said 
that a change occurred 
at time $t$ if $s_{t} > \epsilon_{\lambda}$, where 
$s_{t}$ is a change score,  
and 
$\epsilon_{\lambda}$ was determined according to \eqref{eq:inequality_between_confidence_and_threshold}. 
Type I error probability was defined as the ratio of the trials at which 
an algorithm raises false alarms, 
whereas 
Type II error probability was defined as the ratio of the trials at which 
an algorithm overlooks significant changes: 
\begin{align}
\textrm{Type\, I\, error\, prob.} 
&\mydef 
\frac{ 
  | \left\{ 
  n: 
  s_{t}^{n} \geq \epsilon 
  \mid 
  t \leq 15-\tau, 
  t \geq 15+\tau
  \right\} 
  |
}{ 
  n_{\mathrm{trial}}
},   \\
\textrm{Type\, II\, error\, prob.} 
&\mydef 
\frac{ 
  \left|
    \{
      n: s_{t}^{n} < \epsilon
      \mid 
      15-\tau < t < 15+\tau
    \}
  \right|
}{
  n_{\mathrm{trial}}
}, 
\end{align}
where 
$n_{\mathrm{trial}}$ denotes the number of trials 
for validation. 
$n$ refers to the index of the trials, 
$s_{t}^{n}$ is the change score  
at trial $n$ and time $t$ according to \eqref{eq_MDLCS}, 
and $\tau \in \mathbb{N}$ is an allowed window for false alarms and overlooks. 
We set $n_{\mathrm{trial}}=10$ and $\tau=1$. 

The accuracy 
was also evaluated
using the area under the curve (AUC) score, which is often used 
as a measure of change detection~\cite{Fukushima2019,Fukushima2020}. 
We first fixed the threshold parameter $\epsilon$ 
and converted the change scores $\{ s_{t} \}$ to binary alarms $\{ \alpha_{t} \}$. 
That is, 
$\alpha_{t} = \mathbbm{1}~(s_{t} > \epsilon)$, 
where $\mathbbm{1}(s)$ denotes a binary function that takes on the value of 1 
if and only if $s$ is true. 
We defined $T_{\mathrm{b}}$ as the maximum tolerance delay in change detection. 
We set $T_{\mathrm{b}}=1$. 
When the actual time of the change was $t^{\ast}$, 
and detected change points were 
$\{ t_{k} \}_{k=1}^{m}$, 
we defined the \textit{benefit} of an alarm at time $t_{k}$: 
$b(t_{k}; t^{\ast}) = 1 - |t_{k} - t^{\ast}|/T_{\mathrm{b}}$ \, $(0 \leq |t_{k} - t^{\ast} | < T_{\mathrm{b}})$, 
and $0$ otherwise. 
The number of false alarms was calculated as follows: 
$n_{\mathrm{fa}} = \sum_{k=1}^{m} \alpha_{t_{k}} \mathbbm{1}(b(t_{k}; t^{\ast})) = 0$. 
The AUC 
was 
calculated with recall rate of the total benefit $b/\sup_{\epsilon} b$ and false alarm rate $n_{\mathrm{fa}}/\sup_{\epsilon} n_{\mathrm{fa}}$, 
with $\epsilon$ varying.

For the compression rate of summary graphs, 
we calculated the average code-lengths 
at $t=15$.  

\subsubsection{Results} 
Table~\ref{table:type_I_and_II_error_probabilities_for_the_synthetic_dataset} 
lists the estimated Type I and II error probabilities, AUCs, and  
the code-lengths of the summary graphs.  
Type I and 
II error probabilities for BSC were lower than $0.1$.  
Type I error probabilities of TimeCrunch and DSSG were higher than that of BSC, 
while Type II ones 
were kept relatively low. 
This is because TimeCrunch and DSSG 
detected more subtle changes 
in 
edges 
even when 
significant changes 
did not occur. 
The performance of BSC was superior 
or 
comparable to those of 
the change detection algorithms. 
\begin{table}[tb]
\centering
\caption{Type I and II error probabilities, 
AUCs, 
and code-lengths  
on the synthetic dataset.
}
\label{table:type_I_and_II_error_probabilities_for_the_synthetic_dataset}
\begin{footnotesize}
{\tabcolsep=0.5\tabcolsep
 \renewcommand{\arraystretch}{0.5}
\begin{tabular}{lrrrr}
\toprule
\multicolumn{1}{c}{} & 
\multicolumn{1}{c}{Type I} &  
\multicolumn{1}{c}{Type II} &
\multicolumn{1}{c}{AUC} &
\multicolumn{1}{c}{Code-length} \\ 
\midrule
\multicolumn{1}{l}{BSC ($\lambda=0.5$)} 
& $\mathbf{0.0}$ 
& $0.1$ 
& $0.95 \pm 0.00$
& $144.39 \pm \;\;\;\;\;\;\;\;\;\;\;\;\;\;\;\;\;\; 1.28$
\\
\multicolumn{1}{l}{BSC ($\lambda=1$)}
& $\mathbf{0.0}$ 
& $\mathbf{0.0}$ 
& $\mathbf{1.00 \pm 0.00}$
& $141.82 \pm \;\;\;\;\;\;\;\;\;\;\;\;\;\;\;\;\;\; 1.20$
\\
\multicolumn{1}{l}{BSC ($\lambda=5$)}
& $0.1$ 
& $\mathbf{0.0}$ 
& $0.95 \pm 0.00$
& $\;\;\;\;\;\mathbf{138.53 \pm \;\;\;\;\;\;\;\;\;\;\;\;\;\;\; 1.20}$ 
\\
\midrule 
\multicolumn{1}{l}{TimeCrunch ($m=50$)}
& $0.2$
& $0.1$
& $0.83 \pm 0.02$
& $4101.18 \pm 1656.38$
\\
\multicolumn{1}{l}{TimeCrunch ($m=10$)} 
& $0.2$ 
& $0.2$ 
& $0.78 \pm 0.05$
& $4491.39 \pm 1711.23$
\\
\midrule
\multicolumn{1}{l}{DSSG ($q=0.1$)} 
& $0.3$ 
& $0.6$
& $0.57 \pm 0.00$
& $2656.54 \pm 1006.08$
\\
\multicolumn{1}{l}{DSSG ($q=0.5$)}
& $0.5$
& $0.4$
& $0.56 \pm 0.00$
& $1955.48 \pm \;\;\;\;\;\;872.90$
\\
\midrule
\multicolumn{1}{l}{DeltaCon} 
& $\mathbf{0.0}$
& $0.1$
& $0.95 \pm 0.00$ 
& \multicolumn{1}{c}{---}
\\
\midrule
\multicolumn{1}{l}{Eigenspace-based} 
& $\mathbf{0.0}$ 
& $0.1$ 
& $0.95 \pm 0.00$ 
& \multicolumn{1}{c}{---}
\\
\bottomrule
\end{tabular}
}
\end{footnotesize}
\end{table}
The code-lengths 
with BSC were shorter than those with
TimeCrunch and DSSG. 
This is because of the nature of BSC 
that it encodes 
the original graph structures compactly 
with the hierarchical latent probabilistic variable model. 

\subsection{TwitterWorldCup2014 Dataset}
\label{subsection:twitterworldcup2014_dataset}

We also demonstrated the effectiveness of BSC 
on the TwitterWorldCup2014 dataset\footnote{\url{http://odds.cs.stonybrook.edu/twitterworldcup2014-dataset/}}~\cite{Rayana2016}. 
We constructed entity-entity co-mention 
graphs 
on an hourly basis between twitter hashtags 
from June 1 to July 15~(1,080 time points). 
The total number of entities is 15,856. 

We evaluated BSC, 
DSSG~\cite{Kapoor2020},  
DeltaCon~\cite{Koutra2016}, 
and 
Eigenspace-based algorithm~\cite{Ide2004}  
in terms of 
AUC 
with the maximum tolerance delay $T_{b}=1$ hour.  
For BSC, 
we set 
$k \in \{15$, $20$, $25$, $30$, $35 \}$ 
and 
$\lambda \in \{1$, $100$, $1000\}$. 
We searched 
$q \in \{ 10^{-3}$, $10^{-5} \}$ 
for DSSG. 
We set $w=2$ for the Eigenspace-based algorithm. 
For evaluation, 
the 22 events annotated as ``High importance events’’ were selected. 

Table~\ref{table:aucs_codelengths_and_computation_times_for_the_twitterworldcup2014_dataset} 
summarizes 
the performances 
for the remaining 
games but the first two ones. 
The code-lengths were calculated 
when the events happened in the games, 
and then they are normalized compared to the ones 
with $\lambda=100$. 
BSC was superior to 
other change detection algorithms,  
and 
compressed the original graphs more compactly than DSSG 
with 
much less time. 
\begin{table}[tb]
\centering
\caption{AUCs, relative code-lengths, and computation times 
on the TwitterWorldCup2014 dataset. 
} 
\vspace{-0.3cm}
\label{table:aucs_codelengths_and_computation_times_for_the_twitterworldcup2014_dataset}
\begin{footnotesize}
{\tabcolsep=0.5\tabcolsep
 \renewcommand{\arraystretch}{0.5}
\begin{tabular}{lrrr}
\toprule
\multicolumn{1}{c}{} & 
\multicolumn{1}{c}{AUC} & 
\multicolumn{1}{c}{Relative Code-length} &
\multicolumn{1}{c}{Computation Time~(s)} \\ 
\midrule
\multicolumn{1}{l}{BSC ($\lambda=1$)}
& $0.73$ 
& $1.08 \pm 0.08$ 
& $873$
\\
\multicolumn{1}{l}{BSC ($\lambda=100$)}
& $\mathbf{0.74}$ 
& $1.00 \pm 0.00$ 
& $868$
\\
\multicolumn{1}{l}{BSC ($\lambda=1000$)}
& $0.72$ 
& $\mathbf{0.93 \pm 0.03}$ 
& $874$
\\
\midrule 
\multicolumn{1}{l}{DSSG ($q=10^{-3}$)} 
& $0.53$ 
& $10.23 \pm 0.21$
& $187,497$
\\
\multicolumn{1}{l}{DSSG ($q=10^{-5}$)}
& $0.51$
& $6.76 \pm 0.13$
& $167,322$
\\
\midrule
\multicolumn{1}{l}{DeltaCon} 
& $0.68$ 
& \multicolumn{1}{c}{---}
& $\mathbf{288}$ 
\\
\midrule
\multicolumn{1}{l}{Eigenspace-based} 
& $0.64$ 
& \multicolumn{1}{c}{---} 
& $5,288$
\\
\bottomrule
\end{tabular}
}
\end{footnotesize}
\end{table}

\section{Conclusion}
\label{section:conclusion}

We proposed a new formulation 
with a novel algorithm called 
BSC to balance graph summarization 
and 
change detection 
in a graph stream. 
We introduced the hierarchical latent variable model and efficiently encoded its probabilistic structure with MDL. 
We also theoretically derived the relationship between two parameters 
in hypothesis testing: 
the balancing parameter 
for reduction of superedges in summary graphs and threshold parameter 
for 
change detection in summary graphs. 
Through this relation,  
the latter was determined by the former. 
Experimental results demonstrated the effectiveness of BSC. 

\section*{Acknowledgement}

This work was partially supported by JST KAKENHI 19H01114.

\bibliographystyle{IEEEtran}
\bibliography{bsc}

\end{document}